\documentclass[letterpaper,leqno]{article}
\usepackage{amsmath}
\usepackage{amssymb}
\usepackage{amsthm}
\usepackage[usenames,dvipsnames]{color}
\usepackage{bm}
\usepackage{graphicx}
\usepackage{url}
\usepackage{multirow}
\usepackage{rotating}
\usepackage{wrapfig}
\usepackage[numbers]{natbib}
\usepackage{mdwlist}
\usepackage{algorithm}
\usepackage{algorithmic}
\usepackage{booktabs}
\usepackage[caption=false,font=footnotesize]{subfig}

\usepackage{nips12submit_e}
\graphicspath{
{figures/}
}

\renewcommand{\vec}[1]{\mbox{\boldmath$#1$}}

\newcommand{\trans}[2][]{#2^{#1\!\top}}
\newcommand{\mm}[1]{\mathbf{#1}}

\newcommand{\algorithmicbreak}{\textbf{break}}
\newcommand{\BREAK}{\STATE \algorithmicbreak}

\newtheorem{theorem}{Theorem}

\newtheorem{observation}[theorem]{Observation}

\begin{document}

\nipsfinalcopy
\title{Block Coordinate Descent for Sparse NMF}
\author{ Vamsi K. Potluru \\
Department of Computer Science,\\
University of New Mexico\\
\texttt{ismav@cs.unm.edu}
\And Sergey M. Plis \\
Mind Research Network, \\
\texttt{splis@mrn.org}
\And Jonathan Le Roux  \\
Mitsubishi Electric Research Labs\\
\texttt{leroux@merl.com}
\And
Barak A. Pearlmutter \\
Department of Computer Science,\\
National University of Ireland Maynooth\\
\texttt{barak@cs.nuim.ie}
\And Vince D. Calhoun  \\
Electrical and Computer Engineering, UNM and \\
Mind Research Network\\
\texttt{vcalhoun@mrn.org}
\And Thomas P. Hayes \\
Department of Computer Science,\\
University of New Mexico\\
\texttt{hayes@cs.unm.edu}
}
\maketitle
\begin{abstract}
Nonnegative matrix factorization (NMF) has become a ubiquitous tool for
data analysis.  An important variant is the sparse NMF
problem which arises when we explicitly require the learnt 
features to be sparse. A natural measure of sparsity  is the L$_0$ norm,
however its optimization is NP-hard. 
Mixed norms, such as L$_1$/L$_2$  measure, have been shown to 
model sparsity robustly, based on intuitive attributes that such
measures need to satisfy. This is in contrast to computationally cheaper 
alternatives such as the plain L$_1$ norm.
However, present algorithms designed for optimizing the mixed
norm L$_1$/L$_2$ are slow and other formulations for sparse NMF
have been proposed such as those based on L$_1$ and L$_0$ norms.
Our proposed algorithm allows us to solve the mixed norm sparsity
constraints while not sacrificing computation time.
We present experimental evidence on real-world datasets that shows 
our new algorithm performs an order of magnitude faster compared to the
current state-of-the-art solvers optimizing the mixed norm and is
suitable for large-scale datasets.
\end{abstract}

\section{Introduction}

\begin{figure*}[ht!]
 \centering
 \includegraphics[width=0.53\linewidth]{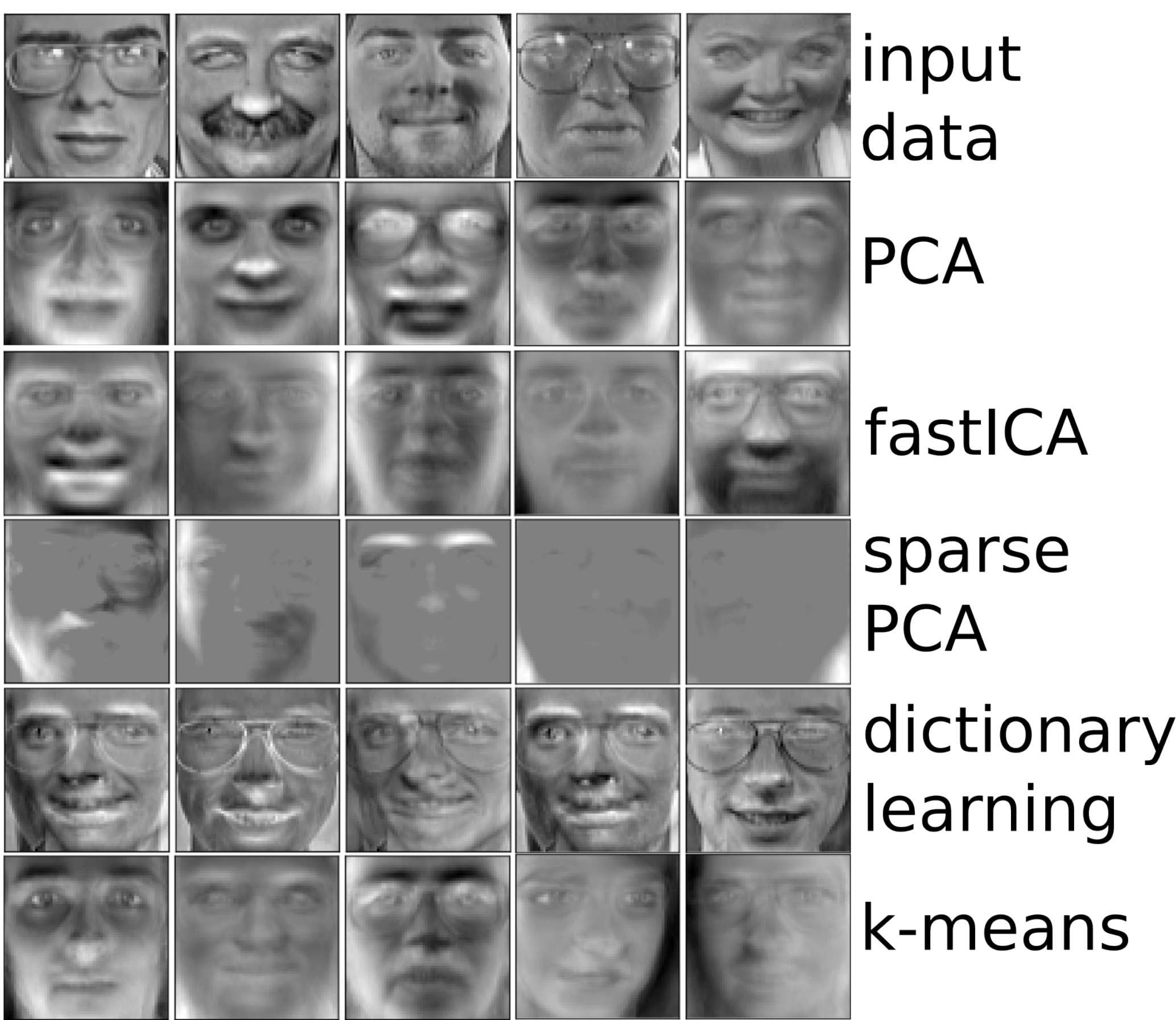}
 \hfill
 \includegraphics[width=0.37\linewidth]{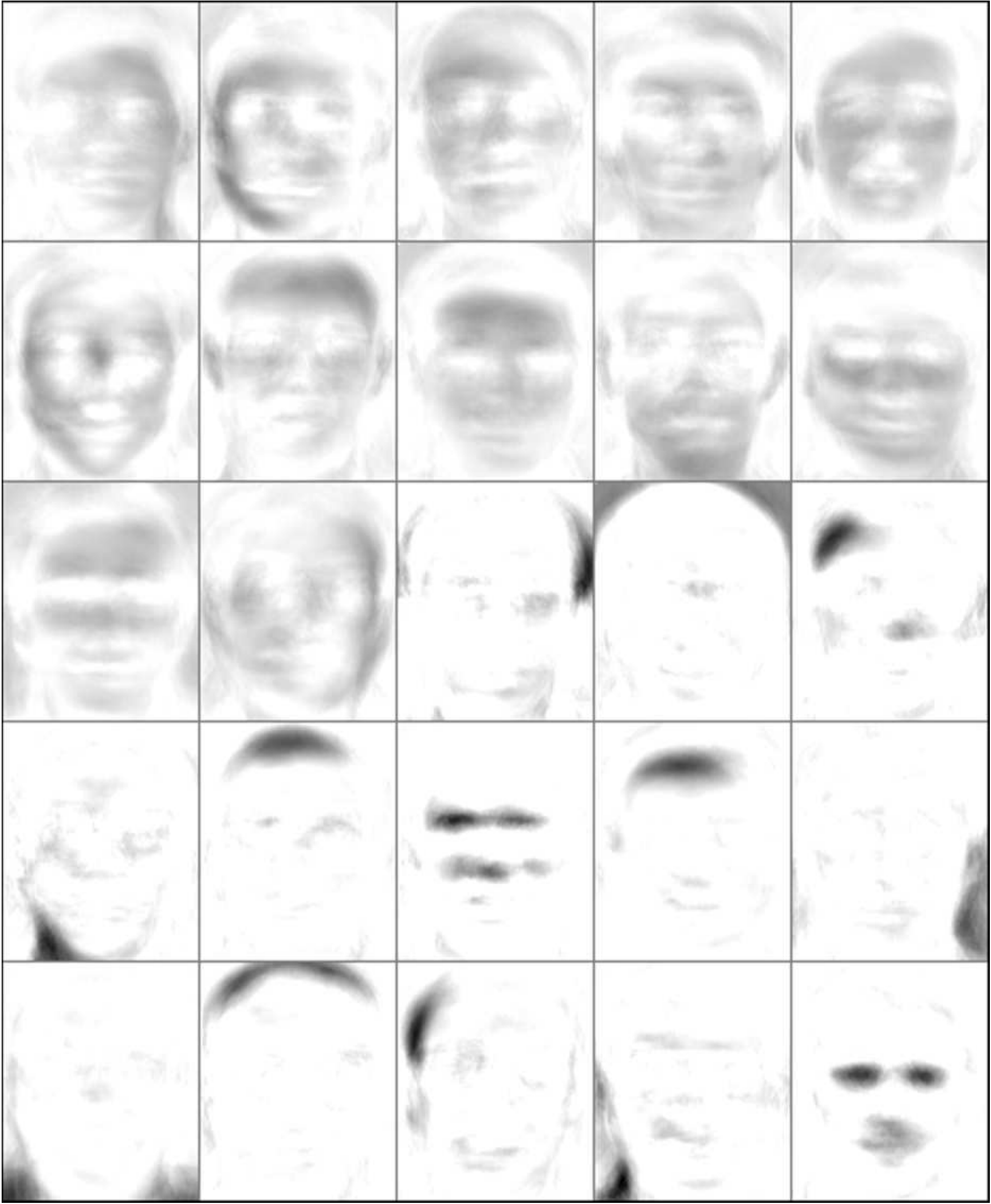}
\caption{(Left) Features learned from the ORL dataset\footnotemark with various matrix 
factorization methods such as principal component analysis (PCA), independent
component analysis (ICA), and dictionary learning. The relative merit of the various
matrix factorizations depends on both the signal domain and the target application
of interest. (Right) Features learned under the sparse NMF formulation where roughly 
half the features were constrained to lie in the interval $[0.2,0.4]$ and the
rest are fixed to sparsity value $0.7$. This illustrates the flexibility
that the user has in fine tuning the feature sparsity based on prior domain knowledge.
White pixels in this figure correspond to the zeros in the features.}
\label{fig:orl_intvl}
\end{figure*}

Matrix factorization arises in a wide range of application domains
and is useful for extracting the latent features in the dataset 
(Figure~\ref{fig:orl_intvl}). 
In particular, we are interested in matrix factorizations which
impose the following requirements:
\begin{itemize}
\item nonnegativity
\item low-rankedness
\item sparsity
\end{itemize}
Nonnegativity is a natural constraint when modeling data with
physical constraints such as chemical concentrations in solutions,
pixel intensities in images and radiation dosages for cancer
treatment. Low-rankedness is useful for learning a lower 
dimensionality representation.  Sparsity is useful for modeling the
conciseness of the representation or that of the latent features. Imposing 
all these requirements on our matrix factorization leads to the sparse 
nonnegative matrix factorization (SNMF) problem.

SNMF    enjoys       quite        a        few
formulations~\cite{berry2007,Hoyer04,Hoyer2002,Heiler06,
Morupl0,Kim2007,Pascual06,Peharz2011}
with successful applications to single-channel speech 
separation~\cite{Schmidt2006} and
micro-array data analysis~\cite{Kim2007,Pascual06}.

However, algorithms~\cite{Hoyer04,Heiler06} for solving SNMF which
utilize the mixed norm of L$_1$/L$_2$ as their sparsity measure 
are slow and do not scale well to large datasets.  
Thus, we  develop an efficient algorithm to solve this problem and
has the following ingredients:
\vspace{-1mm}
\begin{itemize*}
\item A theoretically efficient projection operator ($O(m\log m)$) to enforce the 
user-defined sparsity where $m$ is the dimensionality of the feature vector 
as opposed to the previous approach~\cite{Hoyer04}.  
\item  Novel sequential  updates which provide the  bulk of  our speedup
 compared to the previously employed batch methods~\cite{Hoyer04,Heiler06}.
\end{itemize*}
\vspace{-3mm}

\footnotetext{Scikit-learn package was used in generating the figure.}

\section{Preliminaries and Previous Work}

In this section, we give an introduction to the nonnegative 
matrix factorization (NMF) and SNMF problems. Also, we discuss 
some widely used algorithms from the literature to solve them. 

Both these problems share the following problem and solution
structure. At a high-level, given a nonnegative matrix $\mm{X}$ of 
size $m \times n$, we want to approximate it
with a product  of two nonnegative matrices $\mm{W},  \mm{H}$ of sizes
$m\times r$ and $r\times n$, respectively:
\begin{align}
\label{eq:nmf}
\mm{X} &\approx \mm{W} \mm{H}.
\end{align}
The nonnegative constraint on matrix $\mm{H}$ makes the representation
a conical combination of features given by the columns of matrix $\mm{W}$.
In particular, NMF can result in sparse representations, or a
parts-based representation, unlike other  factorization techniques
such as  principal component analysis (PCA) and vector quantization (VQ).
A common theme in the algorithms proposed for solving these
problems is the use of alternating updates
to the matrix factors, which is natural because the objective function 
to be minimized is convex in $\mm{W}$ and in $\mm{H}$, separately, but
not in both together. Much effort has been focused on optimizing
the efficiency of the core step of updating one of $\mm{W}, \mm{H}$
while the other stays fixed.

\subsection{Nonnegative Matrix Factorization}
\label{sec:NMF}
Factoring a matrix, all of whose entries are nonnegative, as
a product of two low-rank nonnegative factors is a fundamental
algorithmic  challenge. This has arisen naturally in
diverse areas such as image analysis~\cite{LeeSeung99},
micro-array data analysis~\cite{Kim2007},
document clustering~\cite{Xu2003}, chemometrics~\cite{lawton1971},
 information retrieval~\cite{hofmann2001} and biology applications~\cite{buchsbaum2002}.
For further applications, see the references in the following papers~\cite{Arora2012,cohen1993}.

We will consider the following version of the NMF problem, which measures
the reconstruction error using the Frobenius norm~\cite{LeeSeung2001}:
\begin{align}
\min_{\mm{W},\mm{H}}\frac{1}{2}\|\mm{X} &- \mm{W}\mm{H}\|_F^2 \;
                  \mbox{ s.t. } \mm{W}\ge\mm{0}, \, \mm{H}\ge\mm{0} 
  , \,   \|\bm{W}_j\|_2 =1, \; \forall j\in\{1,\cdots,r\} 
\end{align}
where $\ge$ is element-wise. We use subscripts to denote column elements.
Simple multiplicative updates 
were proposed by~\citeauthor{LeeSeung2001} to solve the NMF problem.
This is attractive for the following reasons:
\begin{itemize}
  \item Unlike additive gradient descent methods, there is no
        arbitrary learning rate parameter that needs to be set.
  \item The nonnegativity constraint is satisfied automatically,
        without any additional projection step.
  \item The objective function converges to a limit point and the values are
  non-increasing across the updates, as shown by 
  \citeauthor{LeeSeung2001}~\cite{LeeSeung2001}. 
\end{itemize}
Algorithm~\ref{alg:leeseung} is an example of the kind of multiplicative
update procedure used, for instance, by \citeauthor{LeeSeung2001}
~\cite{LeeSeung2001}. The algorithm
alternates between updating the matrices $\mm{W}$ and $\mm{H}$ (we have only
shown the updates for $\mm{H}$---those for $\mm{W}$ are analogous).
\begin{algorithm}[h!]
\caption{ $\textrm{nnls-mult}(\mm{X},\mm{W},\mm{H})$}
\label{alg:leeseung}
\begin{algorithmic}[1]
\REPEAT
\STATE $\mm{H} = \mm{H} \odot \frac{\trans{\mm{W}}\mm{X}}{\trans{\mm{W}}\mm{W}\mm{H}}$.
\UNTIL convergence
\STATE Output: Matrix $\mm{H}$.
\end{algorithmic}
\end{algorithm}

Here, $\odot$ indicates  element-wise (Hadamard)  product and  matrix division 
is also element-wise.  
To remove the scaling ambiguity, the norm of columns of matrix $\mm{W}$ are 
set to unity. Also, a small constant, say
$10^{-9}$, is added to the denominator in the updates to avoid division by zero.

Besides multiplicative updates, other algorithms have been proposed to solve
the NMF problem based on projected gradient~\cite{Lin2007}, 
block pivoting~\cite{Kim2008}, sequential constrained optimization~\cite{cichockifast09}
and greedy coordinate-descent~\cite{Hsieh11}.

\subsection{Sparse Nonnegative Matrix Factorization}
\label{sec:sparseNMF}
The     nonnegative     decomposition     is    in     general     not
unique~\cite{NIPS2003_LT10}.  
Furthermore, the  features may not  be parts-based
if the data  resides well inside the  positive orthant.
To address  these issues, sparseness  constraints have been
imposed  on  the  NMF  problem.

Sparse NMF can be formulated in many different ways. From a user point of view,
we can split them into two classes of formulations: explicit and implicit. 
In explicit versions of SNMF~\cite{Hoyer04,Heiler06}, one can set the
sparsities of the matrix factors $\bm{W},\bm{H}$ directly. On the other hand, in implicit versions
of SNMF~\cite{Kim2007,Pascual06}, the sparsity is controlled via a regularization 
parameter and is often hard to tune to specified sparsity values a priori. However, the 
algorithms for implicit versions tend to be faster compared to the explicit versions of 
SNMF.

In this paper, we consider the explicit sparse NMF formulation 
proposed by~\citeauthor{Hoyer04} ~\cite{Hoyer04}. To make the presentation easier 
to follow, we first consider the case where the
sparsity is imposed on one of the matrix factors, namely
the feature matrix $\bm{W}$---the analysis for the symmetric case
where the sparsity is instead set on the other matrix factor $\bm{H}$
is analogous. The case where sparsity requirements are imposed on both the
matrix factors is dealt with in the Appendix. 
The sparse NMF problem formulated by~\citeauthor{Hoyer04}~\cite{Hoyer04} with 
sparsity on matrix $\bm{W}$  is as follows:
\begin{align}
\label{prob:sparse}
\nonumber
  \min_{\mm{W},\mm{H}} f(\bm{W},\bm{H}) =& \frac{1}{2}\|\mm{X}-\mm{W}\mm{H}\|_F^2 
                  \; \mbox{s.t. }  \mm{W}\ge\mm{0},\mm{H}\ge\mm{0}, \\
    &\|\bm{W}_j\|_2 =1, \, \textrm{sp}(\bm{W}_j) =\alpha, \; \forall j\in\{1,\cdots,r\} 
\end{align}
Sparsity measure for a $d$-dimensional vector $\vec{x}$ is
given by:  
\begin{align}
\label{eq:spar}
\textrm{sp}(\vec{x}) =& \frac{\sqrt{d} -\|x\|_1/\|x\|_2}
                                  {\sqrt{d}-1} 
\end{align}
The sparsity measure~\eqref{eq:spar} defined above has many appealing qualities.
Some of which are as follows:
\begin{itemize}
\item The measure closely models the intuitive notion of sparsity
as captured by the $L_0$ norm. So, it easy for the user to
specify sparsity constraints from prior knowledge of the application
domain.
\item Simultaneously, it is able to avoid the pitfalls associated
with directly optimizing the $L_0$ norm. Desirable properties for sparsity
measures have been previously explored~\cite{Hurley2009} and it satisfies
all of these properties for our problem formulation. The properties 
can be briefly summarized as:  (a) Robin Hood --- Spreading the energy from  
larger coordinates to smaller ones decreases sparsity,  
(b) Scaling --- Sparsity is invariant to scaling, (c) Rising tide ---
Adding a constant to the coordinates decreases sparsity, (d) Cloning ---
Sparsity is invariant to cloning, (e) Bill Gates --- One big coordinate can 
increase sparsity, (f) Babies --- coordinates with zeros increase sparsity.
\item The above sparsity measure enables one to limit the sparsity for
each feature to lie in a given range by changing the equality constraints
in the SNMF formulation~\eqref{prob:sparse} to inequality constraints~\cite{Heiler06}.
This could be useful in scenarios like fMRI brain analysis, where one would like
to model the prior knowledge such as sizes of artifacts are different from
that of the brain signals. A sample illustration on a face dataset is shown
in Figure~\ref{fig:orl_intvl} (Right). The features are now evenly split into two
groups of local and global features by choosing two different intervals of sparsity.

\end{itemize}

A gradient descent-based
algorithm called Nonnegative Matrix Factorization with Sparseness Constraints (NMFSC)
to solve SNMF was proposed~\cite{Hoyer04}.
Multiplicative updates were used for optimizing the matrix factor which did not have
sparsity constraints specified.
~\citeauthor{Heiler06}\cite{Heiler06} proposed two new algorithms  which also
solved this problem by sequential cone programming and utilized general purpose 
solvers like MOSEK (\url{http://www.mosek.com}). We will consider the
faster one of these called tangent-plane constraint (TPC) algorithm.
However, both these algorithms, namely NMFSC and TPC, solve for the
whole matrix of coefficients at once. In contrast, we propose a 
block coordinate-descent strategy which considers a sequence of vector 
problems where each one can be solved in closed form efficiently.

\section{The Sequential Sparse NMF Algorithm}
\label{sec:our}
We present our algorithm which we call \textbf{S}equential 
\textbf{S}parse \textbf{NMF} (\textbf{SSNMF}) to solve the SNMF problem
as follows:

First, we consider a problem of special form  
which is the building block (Algorithm~\ref{alg:subroutine})
of our SSNMF algorithm and give an efficient, as well as exact, algorithm to
solve it.  Second, we describe our sequential approach (Algorithm~\ref{alg:spar})
to solve the subproblem of SNMF. This uses the routine we developed in the previous step.
Finally, we combine our routines developed in the previous two steps
along with standard solvers (for instance Algorithm~\ref{alg:leeseung})
to complete the SSNMF Algorithm (Algorithm~\ref{alg:ssnmf}).

\subsection{Sparse-opt}
\label{sec:sparseopt}
Sparse-opt routine solves the following subproblem which arises when solving 
problem~\eqref{prob:sparse}: 
\begin{align}
\label{prob:sparse_sub}
\max_{\vec{y}\ge0} \trans{\vec{b}}\vec{y}  \textrm{  s.t. }  
      &\|\vec{y}\|_1=k , \|\vec{y}\|_2 =1 
      \end{align}
where vector $\vec{b}$ is of size $m$. 
This problem has been previously considered~\cite{Hoyer04}, and an algorithm to solve it 
was proposed which we will henceforth refer to as the Projection-Hoyer.  
Similar projection problems have been recently considered in the literature and 
solved efficiently~\cite{duchi2008,chen2011}.
\begin{observation} 
\label{obs:phase}
    For any $i,j$, we have that if $b_i \ge b_j$, then 
      $y_i \ge y_j$.
\end{observation}
Let us first consider the case when the vector $\vec{b}$ is sorted. Then
by the previous observation, we have a transition point $p$ that separates the 
zeros of the solution vector from the rest. 
\begin{observation} 
By applying the Cauchy-Schwarz inequality on $\vec{y}$ and the all ones vector, 
we get $p\ge k^2$.
\end{observation}
The Lagrangian of the problem~\eqref{prob:sparse_sub} is :
\begin{align}
\nonumber
L(\vec{y},\mu,\lambda,\vec{\gamma})= \trans{\vec{b}}\vec{y} + 
              \mu\left(\sum_{i=1}^m y_i -k\right) 
            &+ \frac{\lambda}{2}\left(\sum_{i=1}^m y_i^2-1\right) 
             + \trans{\vec{\gamma}}\vec{y}
        \end{align}
  Setting the partial derivatives of the Lagrangian to zero, we get
     by observation~\ref{obs:phase}:
   \begin{align*}
       \sum_i^m y_i& = k,  
          \sum_i^m y_i^2 = 1\\
       b_i + \mu(p) + \lambda(p) y_i &=0   , \forall i \in \{1,2,\cdots,p\} \\
         \gamma_i&=0 ,\forall i \in \{1,\cdots,p\} \\
          y_i &=0   , \forall i \in \{p+1,\cdots,m\} 
    \end{align*}        
where we account for the dependence of the Lagrange parameters 
$\lambda,\mu$ on the transition point $p$. We compute the objective value 
of problem~\eqref{prob:sparse_sub} for all transition points $p$ in the 
range from $k^2$  to $m$ and select the one with the highest value.
In the case, where the vector $\vec{b}$ is not sorted, we just simply sort
it and note down the sorting permutation vector. The complete 
algorithm is given in Algorithm~\ref{alg:subroutine}.
The dominant contribution to the running time of Algorithm~\ref{alg:subroutine}
is the sorting of vector $\vec{b}$ and therefore can be implemented in 
$O(m\log m)$ time\footnote{This can be further reduced to linear time by noting that
we do not need to fully sort the input in order to find $p*$.}.  
Contrast this with the running time of Projection-Hoyer whose worst
case is $O(m^2)$~\cite{Hoyer04,theis2005}.

\begin{algorithm}
\caption{ $\textrm{Sparse-opt}(\vec{b},k)$} 
\label{alg:subroutine}
\begin{algorithmic}[1]
\STATE Set $\vec{a}= \textrm{sort}(\vec{b} )$ and $p^*=m$. Get a mapping $\pi$ such 
that $a_i=b_{\pi(i)}$ and $a_j \ge a_{j+1}$ for all valid $i,j$.
\STATE  Compute values of $\mu(p),\lambda(p)$ as follows:
\FOR{$p \in \{\lceil k^2 \rceil$,m\}}
\STATE      $\lambda(p) = -\sqrt{{\frac{p\sum_{i=1}^p a_i^2 -\left(\sum_{i=1}^{p} a_i\right)^2}{\left(p-k^2\right)}}} $
\STATE      $\mu(p) = -\frac{\sum_{i=1}^p a_i}{p} -\frac{k}{p}\lambda(p) $ 
      \IF{$a(p)<-\mu(p)$}
            \STATE   $p^*=p-1$
            \BREAK  
      \ENDIF      
\ENDFOR
\STATE Set $x_i = -\frac{a_i+\mu(p^*)}{\lambda(p^*)} ,\forall i\in\{1,\cdots,p^*\}$ and to
zero otherwise.
\STATE  Output: Solution vector $\vec{y}$ where $y_{\pi(i)}=x_i$.  
\end{algorithmic}
\end{algorithm}


\subsection{Sequential Approach ---Block Coordinate Descent} 
\label{sec:seq-up}
Previous approaches for solving SNMF~\cite{Hoyer04,Heiler06} use 
batch methods to solve for sparsity constraints. That is, the whole matrix
is updated at once and projected to satisfy the constraints. We take a
different approach of updating a column vector at a time. This gives us the benefit
of being able to solve the subproblem (column) efficiently and exactly. Subsequent updates
can benefit from the newly updated columns resulting in faster convergence as seen
in the experiments.

In particular, consider the optimization problem~\eqref{prob:sparse} for a column $j$
of the matrix
$\mm{W}$ while fixing the rest of the elements of matrices $\mm{W},\mm{H}$:
\begin{align}
\nonumber
   \min_{\mm{W}_j\ge\vec{0}} \tilde{f}(\mm{W}_j) & = 
 \frac{1}{2} g \|\mm{W}_j\|^2_2  
          + \trans{\vec{u}}\mm{W}_{j}  \; \textrm{   s.t. } \, \|\mm{W}_j\|_2=1 ,\,
          \|\mm{W}_j\|_1= k 
\end{align}
where $g=\trans{\bm{H}_j}\bm{H}_j$ and $u=-\bm{X}\trans{\bm{H}_j} + \sum_{i\ne j}
\bm{W}_i(\bm{H}\trans{\bm{H}})_{ij}$.
This reduces to the problem~\eqref{prob:sparse_sub}  for which we have proposed an
exact algorithm (Algorithm~\ref{alg:subroutine}). We update
the columns of the matrix factor $\mm{W}$ sequentially as shown in Algorithm~\ref{alg:spar}.
We call it sequential for we update the columns one at a time.
Note that this approach can be seen as an instance of block coordinate descent methods by
mapping features to blocks and the Sparse-opt projection operator to a descent step.
\begin{algorithm}[h!]
\caption{$\textrm{sequential-pass}(\mm{X},\mm{W},\mm{H})$}
\label{alg:spar}
\begin{algorithmic}[1]
\STATE $\mm{C} = -\mm{X}\trans{\mm{H}} + \mm{W}\mm{H}\trans{\mm{H}}$
\STATE $\mm{G} = \mm{H}\trans{\mm{H}}$
\REPEAT
\FOR{ j = $1$ to $r$ (randomly) }
\STATE $\mm{U}_j=\mm{C}_j-\mm{W}_jG_{jj}$ 
\STATE $\vec{t}$ = Sparse-opt($-\mm{U}_j,k$).
\STATE $\mm{C} =  \mm{C} + \left(\vec{t}-\mm{W}_j\right)\trans{\vec{G}}_j$
\STATE $\mm{W}_j = \vec{t}$.
\ENDFOR
\UNTIL convergence
\STATE Output: Matrix $\mm{W}$.
\end{algorithmic}
\end{algorithm}

\subsection{SSNMF Algorithm for Sparse NMF}
We are now in a position to present our complete Sequential Sparse NMF (SSNMF) algorithm. 
By combining Algorithms~\ref{alg:leeseung},~\ref{alg:subroutine} and~\ref{alg:spar}, 
we obtain SSNMF (Algorithm~\ref{alg:ssnmf}). 

\begin{algorithm}[h!]
\caption{$\textrm{ssnmf}(\mm{X},\mm{W},\mm{H})$}
\label{alg:ssnmf}
\begin{algorithmic}[1]
\REPEAT
\STATE $\mm{W}=\textrm{sequential-pass}(\mm{X},\mm{W},\mm{H})$
\STATE $\mm{H}=\textrm{nnls-mult}(\mm{X},\mm{W},\mm{H})$
\UNTIL{convergence}
\STATE Output: Matrices $\mm{W},\mm{H}$. 
\end{algorithmic}
\end{algorithm}

\section{Implementation Issues}
\label{sec:implement}
For clarity of exposition, we presented the plain vanilla version of
our SSNMF Algorithm~\ref{alg:ssnmf}. We now describe some of the actual
implementation details.  
\begin{itemize}
\item{Initialization:}
Generate a positive random vector $\vec{v}$ of size $m$ and obtain 
$\vec{z}=\textrm{Sparse-opt}(\vec{v},k)$ where $k=\sqrt{m}-\alpha\sqrt{m-1}$
(from equation~\eqref{eq:spar}). 
Use the solution $\vec{z}$ and its random permutations to initialize matrix $\mm{W}$. 
Initialize the matrix $\mm{H}$ to uniform random entries in $[0,1]$.

\item{Incorporating faster solvers:}
We use multiplicative updates for a fair comparison with 
NMFSC and TPC. However, we can use other NNLS solvers 
~\cite{Lin2007,Kim2008,cichockifast09,Hsieh11} to solve for
matrix $\bm{H}$. Empirical results (not reported here) show that
this further speeds up the SSNMF algorithm.

\item{Termination:}
In our experiments, we fix the number of alternate updates or 
equivalently the number of times we update matrix $\bm{W}$. 
Other approaches include specifying total running time, relative
change in objective value between iterations or approximate
satisfaction of KKT conditions.

\item{Sparsity constraints:}
We have primarly considered the sparse NMF model as formulated
by~\citeauthor{Hoyer04}~\cite{Hoyer04}. This has been generalized
by~\citeauthor{Heiler06}~\cite{Heiler06} by relaxing the sparsity constraints to lie in
user-defined intervals. Note that, we can handle this 
formulation~\cite{Heiler06} by making a trivial change to
Algorithm~\ref{alg:spar}.

\end{itemize}

\section{Experiments and Discussion}
\label{sec:exp}
In this section, we compare the performance of our
algorithm with the state-of-the-art NMFSC and TPC algorithms
~\cite{Hoyer04,Heiler06}. Running times for the algorithms
are presented when applied to one synthetic and three real-world
datasets.
Experiments report reconstruction error ($\|\mm{X}-\mm{W}\mm{H}\|_F$)
instead of objective value for convenience of display.
For all experiments on the datasets, we ensure that our final
reconstruction error is always better than that of the other two algorithms.
Our algorithm was implemented in MATLAB (\url{http://www.mathworks.com}) similar to NMFSC and TPC.
All of our experiments were run on a $3.2$Ghz Intel machine with $24$GB of
RAM and the number of threads set to one.

\subsection{Datasets}
For comparing the performance of SSNMF with NMFSC and TPC, we consider
the following synthetic and three real-world datasets : 
\begin{itemize}
\item Synthetic: $200$ images of size $9 \times 9$ as provided
by ~\citeauthor{Heiler06}~\cite{Heiler06} in their code implementation.
\item CBCL face dataset consists of $2429$ images of size $19 \times 19$ and can be obtained at
\url{http://cbcl.mit.edu/cbcl/software-datasets/FaceData2.html}.
\item ORL: Face dataset that consists of $400$ images of size $112
\times 92$ and can be obtained at \url{http://www.cl.cam.ac.uk/research/dtg/attarchive/facedatabase.html}.
\url{http://www.kyb.tuebingen.mpg.de/bethge/vanhateren/iml/}.
We use $400$ of these images in our experiments.
\item sMRI:
Structural MRI scans of $269$ subjects taken at
the John Hopkins University were obtained. The scans were
taken on a single 1.5T scanner with the imaging
parameters set to $35$mm TR, $5$ms TE, matrix size of $256 \times 256$. 
We segment these images 
into gray matter, white matter and cerebral spinal fluid
images, using the software program SPM5
(http://www.fil.ion.ucl.ac.uk/spm/software/spm5/),
followed by spatial smoothing with a Gaussian kernel
of $10 \times 10 \times 10$ mm. This results in images which are
of size $105 \times 127 \times 46$
\end{itemize}

\begin{figure*}[ht!]
 \centering
   \subfloat{\includegraphics[width=0.9\linewidth]{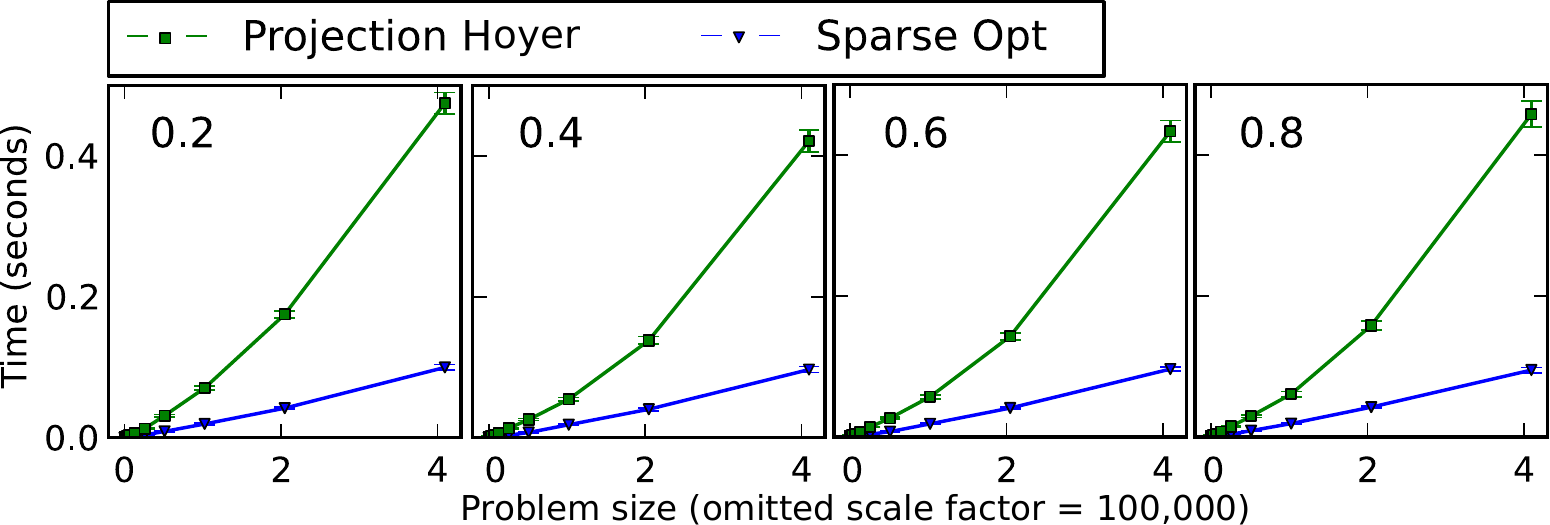}}
   \hfill
\caption{Mean running times for Sparse-opt and the  
Projection-Hoyer are presented for random problems. The x-axis
plots the dimension of the problem while
the y-axis has the running time in seconds. Each of the subfigures corresponds to
a single sparsity value in $\{0.2,0.4,0.6,0.8\}$. Each datapoint corresponds to the 
mean running time averaged over $40$ runs for random problems of the same fixed dimension.
}
\label{fig:comp_our_hoy}
\end{figure*}

\begin{figure*}[ht!]
\includegraphics[width=1\linewidth]{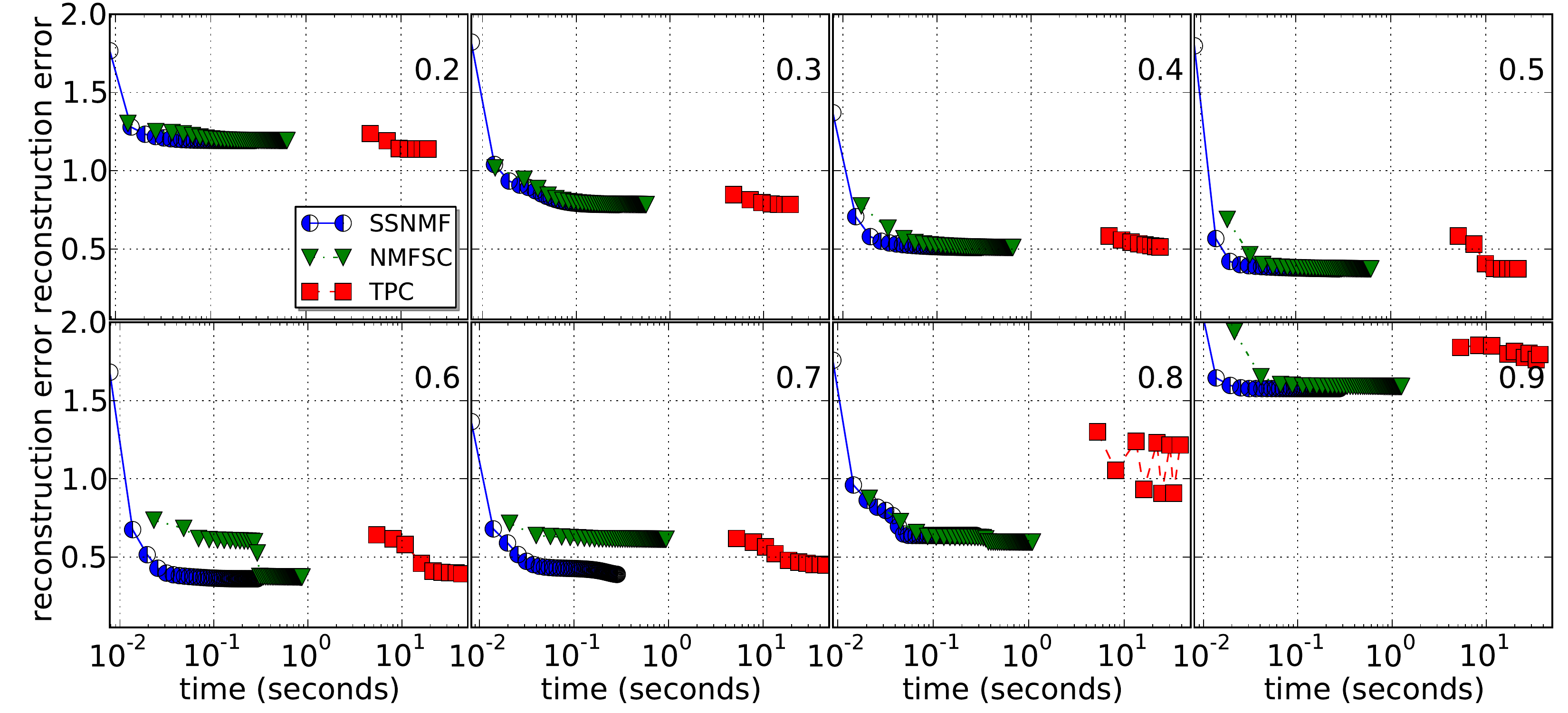}
\caption{Running times for SSNMF  and NMFSC and TPC algorithms
on the synthetic dataset where the sparsity values range from
$0.2$ to $0.8$ and number of features is $5$. Note that SSNMF 
and NMFSC are over an order of magnitude faster than TPC.}
\label{fig:all}
\end{figure*}

\begin{figure*}[ht!]
\includegraphics[width=1\linewidth]{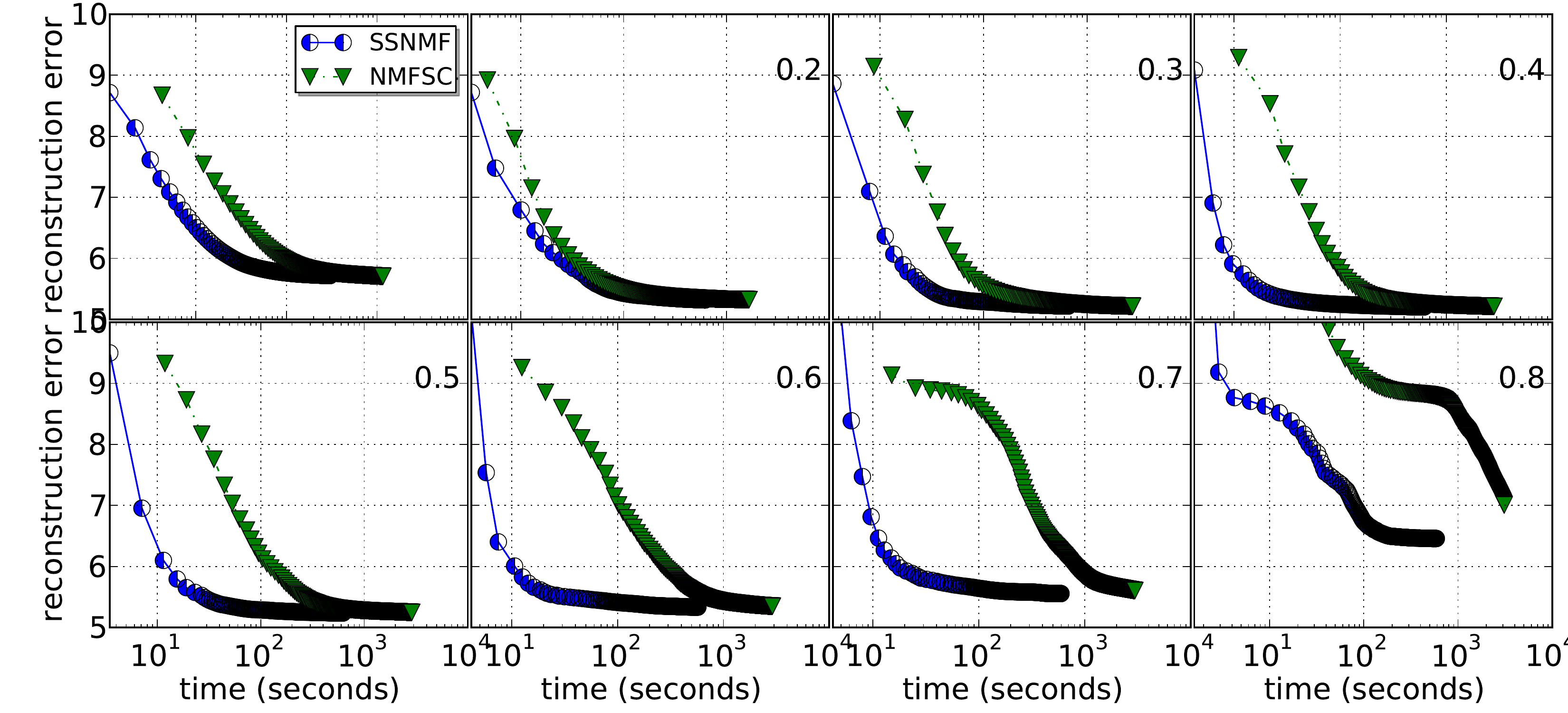}
\caption{Convergence plots for the ORL dataset with sparsity from $[0.1,0.8]$ 
for the NMFSC and SSNMF algorithms.
Note that we are an order of magnitude faster, especially when the sparsity is higher.} 
\label{fig:ORL-speedup}
\end{figure*}

\begin{figure*}[ht!]
\includegraphics[width=1\linewidth]{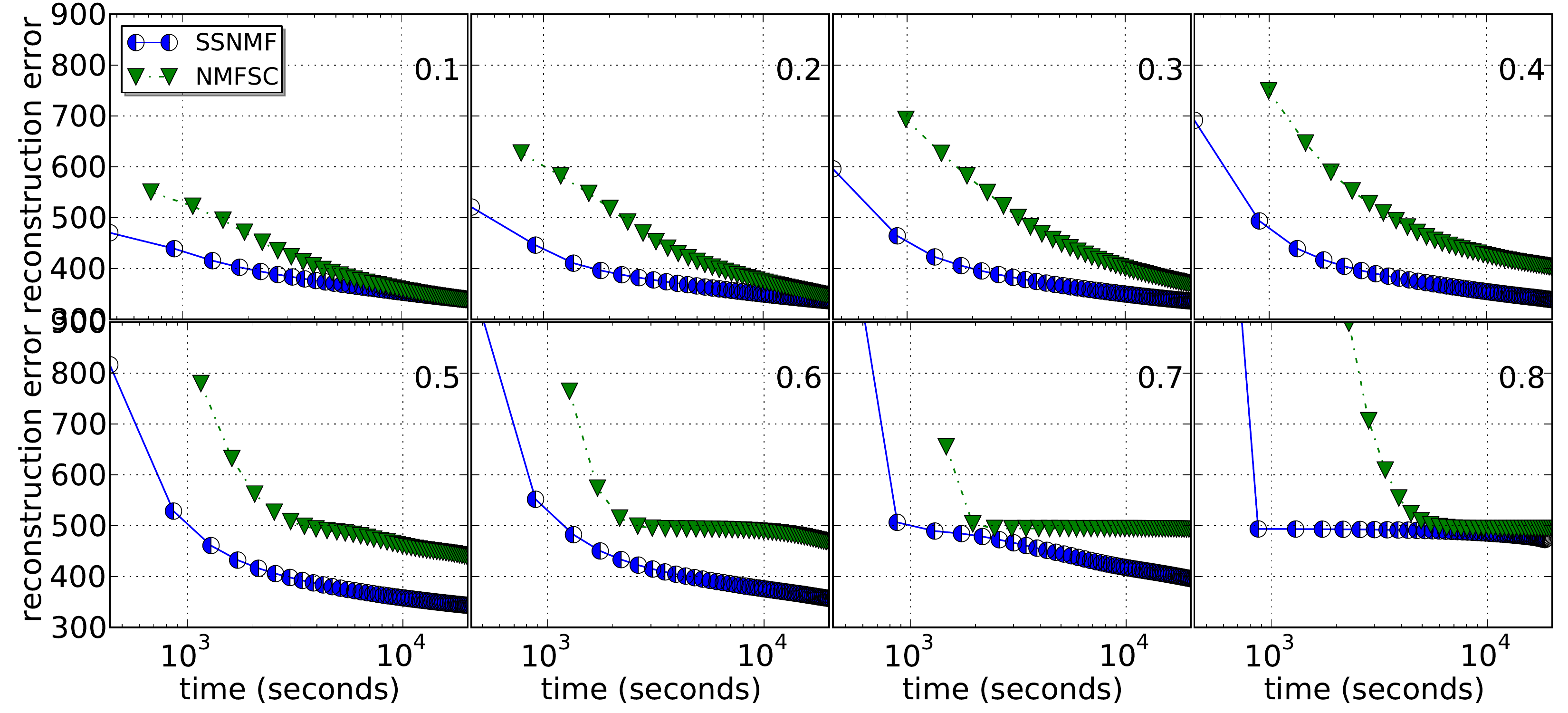}
\caption{Running times for SSNMF and NMFSC algorithms
for the sMRI dataset with rank set to $40$ and
sparsity values of $\alpha$ from $0.1$ to $0.8$. Note that for higher sparsity
values we converged to a lower reconstruction error and are also
noticeably faster than the NMFSC algorithm.}
\label{fig:smri_comp}
\end{figure*}

\subsection{Comparing Performances of Core Updates } 
We compare our Sparse-opt (Algorithm~\ref{alg:subroutine}) routine 
with the competing Projection-Hoyer~\cite{Hoyer04}.
In particular, we generate $40$ random problems for each sparsity constraint
in $\{0.2,0.4,0.6,0.8\}$ and a fixed problem size.
The problems are of size $2^i\times100$ where $i$ takes integer values from $0$ to $12$.  
Input coefficients are generated by drawing samples uniformly at random from $[0,1]$.
The mean values of the running times for Sparse-opt and the Projection-Hoyer for
each dimension and corresponding sparsity value are plotted in 
Figure~\ref{fig:comp_our_hoy}. 

\begin{figure*}[ht!]
\includegraphics[width=1\linewidth]{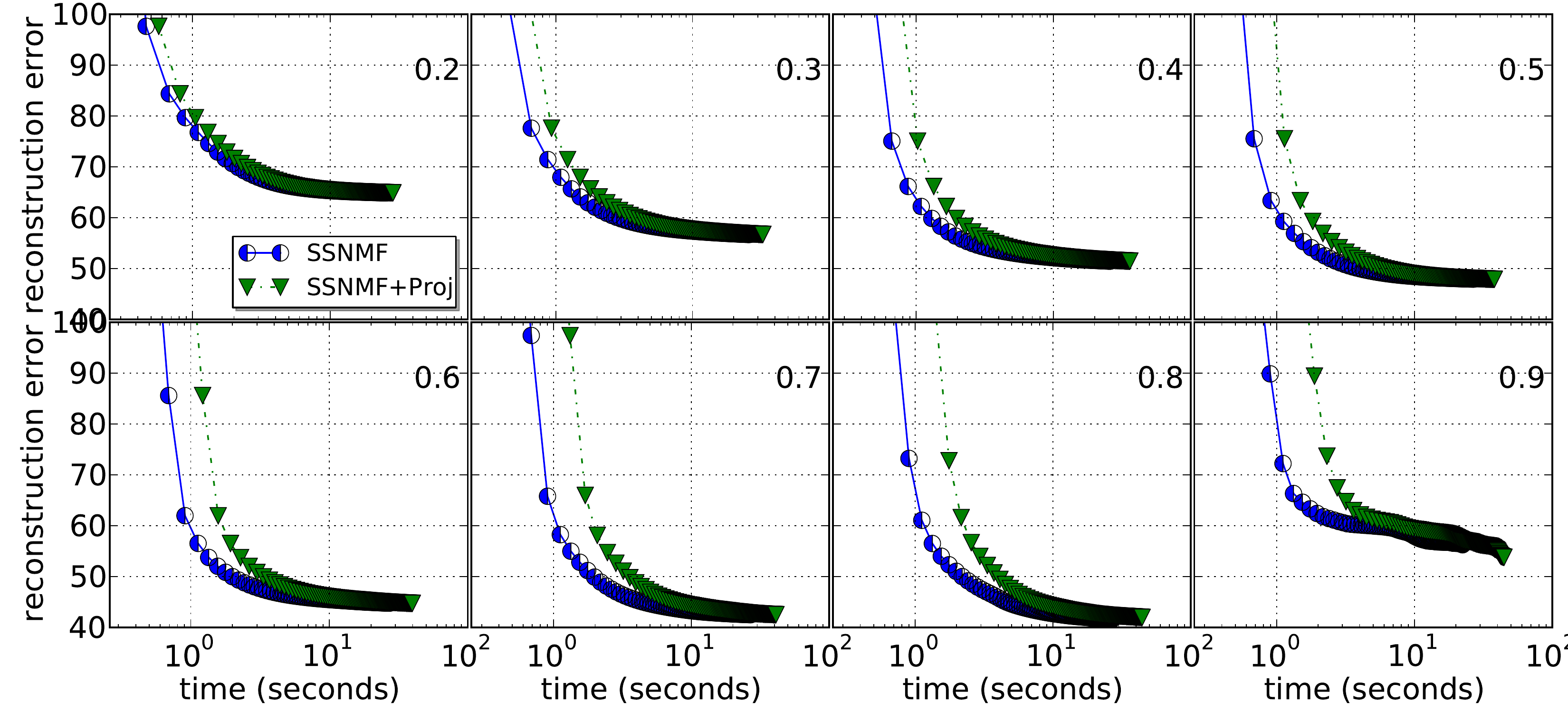}
\caption{Running times for SSNMF  and SSNMF+Proj algorithms
for the CBCL face dataset with rank set to $49$ and
sparsity values ranging from $0.2$ to $0.9$}
\label{fig:cbcl_comp_proj}
\end{figure*}

We compare SSNMF with SSNMF+Proj on the CBCL dataset.
The algorithms were run with rank set to $49$.
The running times are shown in Figure~\ref{fig:cbcl_comp_proj}.
We see that in low-dimensional datasets, the difference in
running times are very small.

\subsection{Comparing Overall Performances}
\paragraph{SSNMF versus NMFSC and TPC:}
We plot the performance of SSNMF against NMFSC and TPC on the
synthetic dataset provided by~\citeauthor{Heiler06}~\cite{Heiler06} in 
Figure~\ref{fig:all}. We used the
default settings for both  NMFSC and TPC using the software
provided by the authors. Our experience with TPC was not encouraging
on bigger datasets and hence we show its performance only on 
the synthetic dataset. It is possible that the performance of TPC can be 
improved by changing the default settings but we found it 
non-trivial to do so.

\paragraph{SSNMF versus NMFSC:}
To ensure fairness, we removed logging information from NMFSC code~\cite{Hoyer04} and only 
computed the objective for equivalent number of matrix updates
as SSNMF. We do not plot the objective values at the first iteration
for convenience of display. However, they are the same for both algorithms
because of the shared initialization .
We ran the SSNMF and NMFSC on the ORL face dataset.
The rank was fixed at $25$ in both the algorithms. 
Also, the plots of running times versus objective values are shown 
in Figure~\ref{fig:ORL-speedup} corresponding to sparsity values ranging from $0.1$ to $0.7$.
Additionally, we ran our SSNMF algorithm and NMFSC algorithm on a large-scale dataset 
consisting of the structural MRI images by setting the rank to $40$. The running 
times are shown in Figure~\ref{fig:smri_comp}.

\subsection{Main Results}
We compared the running times of our Sparse-opt routine versus the
Projection-Hoyer and found that on the synthetically generated datasets
we are faster on average.

Our results on switching the Sparse-opt
routine with the Projection-Hoyer did not slow down our SSNMF 
solver significantly for the datasets we considered. So, we conclude
that the speedup is mainly due to the sequential nature of the
updates (Algorithm~\ref{alg:spar}).

Also, we converge faster than NMFSC for fewer number of
matrix updates. This can be seen by noting that the plotted points
in Figures~\ref{fig:ORL-speedup} and~\ref{fig:smri_comp}  are such
that the number of matrix updates are the same for both SSNMF and NMFSC.
For some datasets, we noted a speedup of an order of magnitude making 
our approach attractive for computation purposes.

\begin{figure*}[ht!]
 \centering
   \subfloat[sparsity 0.5]{{\includegraphics[width=0.15\linewidth]
   {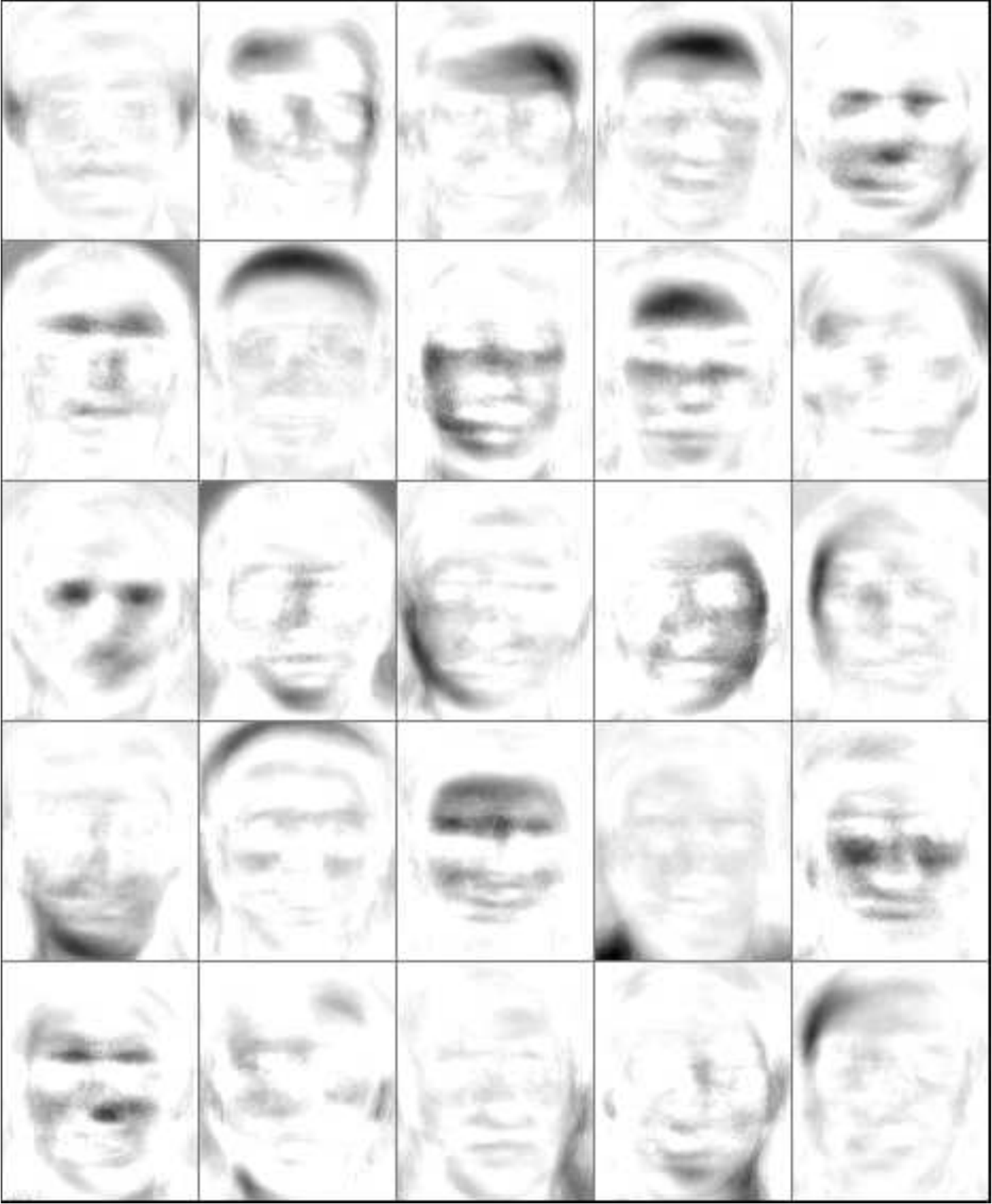} }
  {\includegraphics[width=0.15\linewidth]{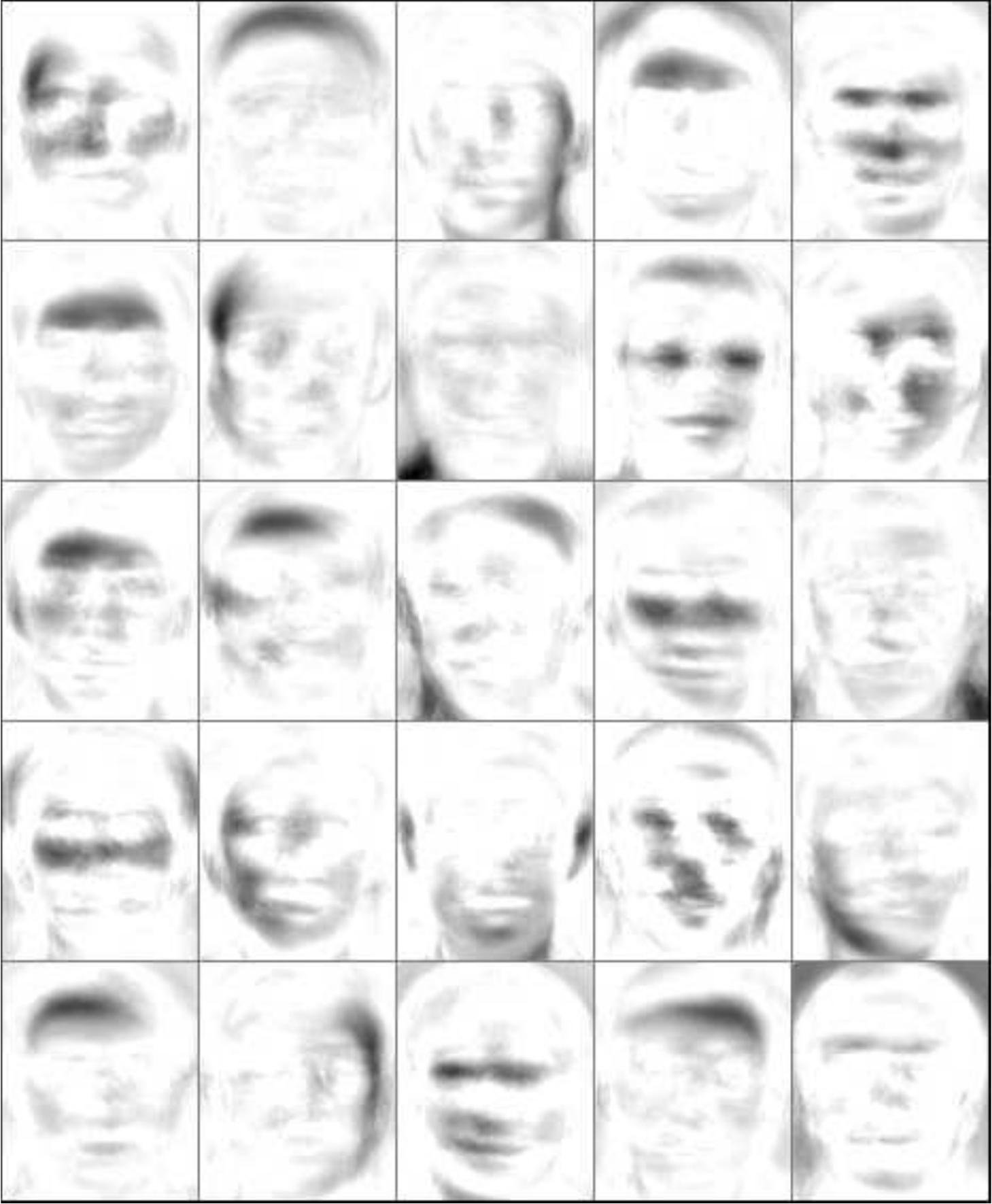}}} 
   \hfill
   \subfloat[sparsity 0.6]{{\includegraphics[width=0.15\linewidth]
   {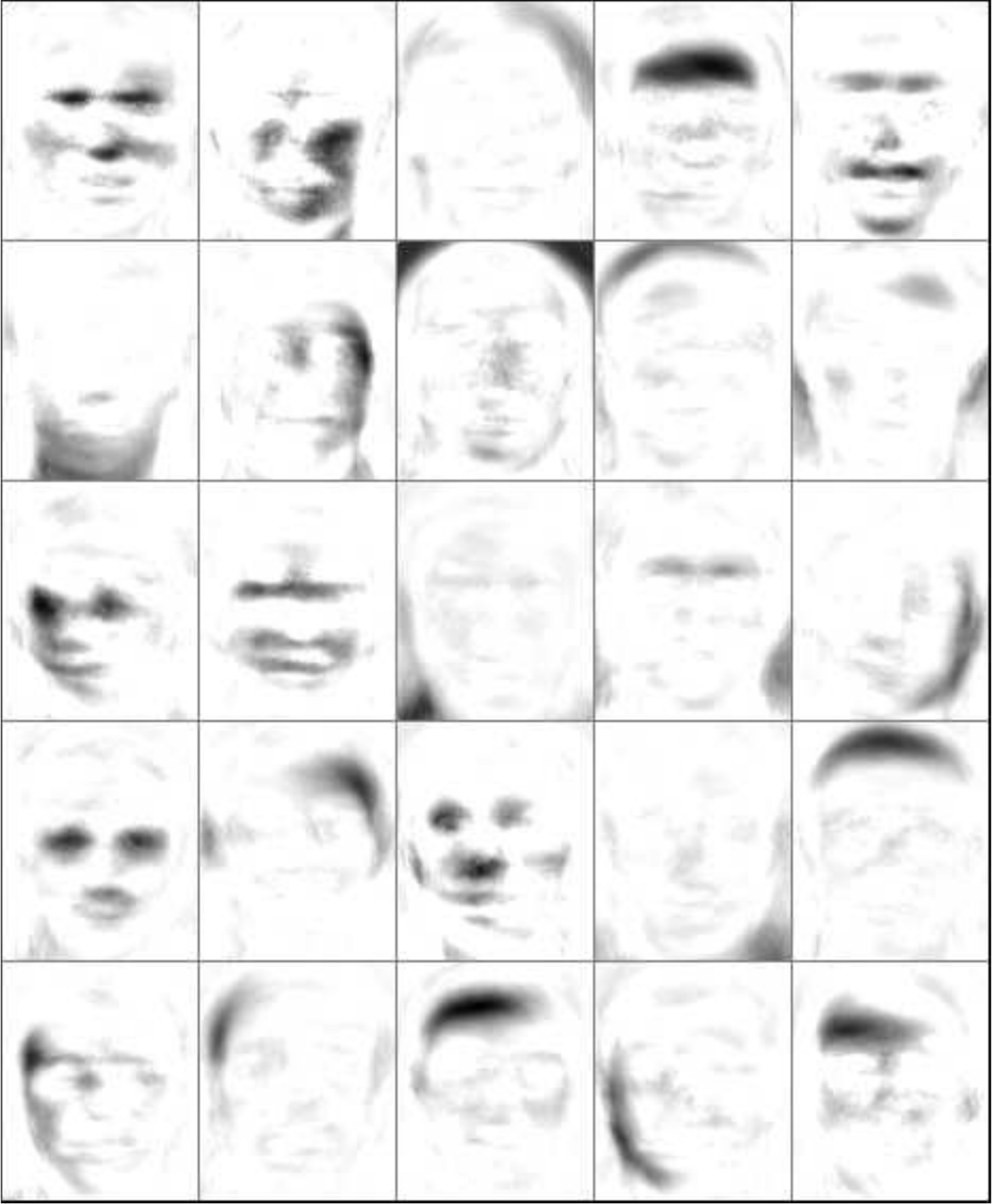}}
    {\includegraphics[width=0.15\linewidth]{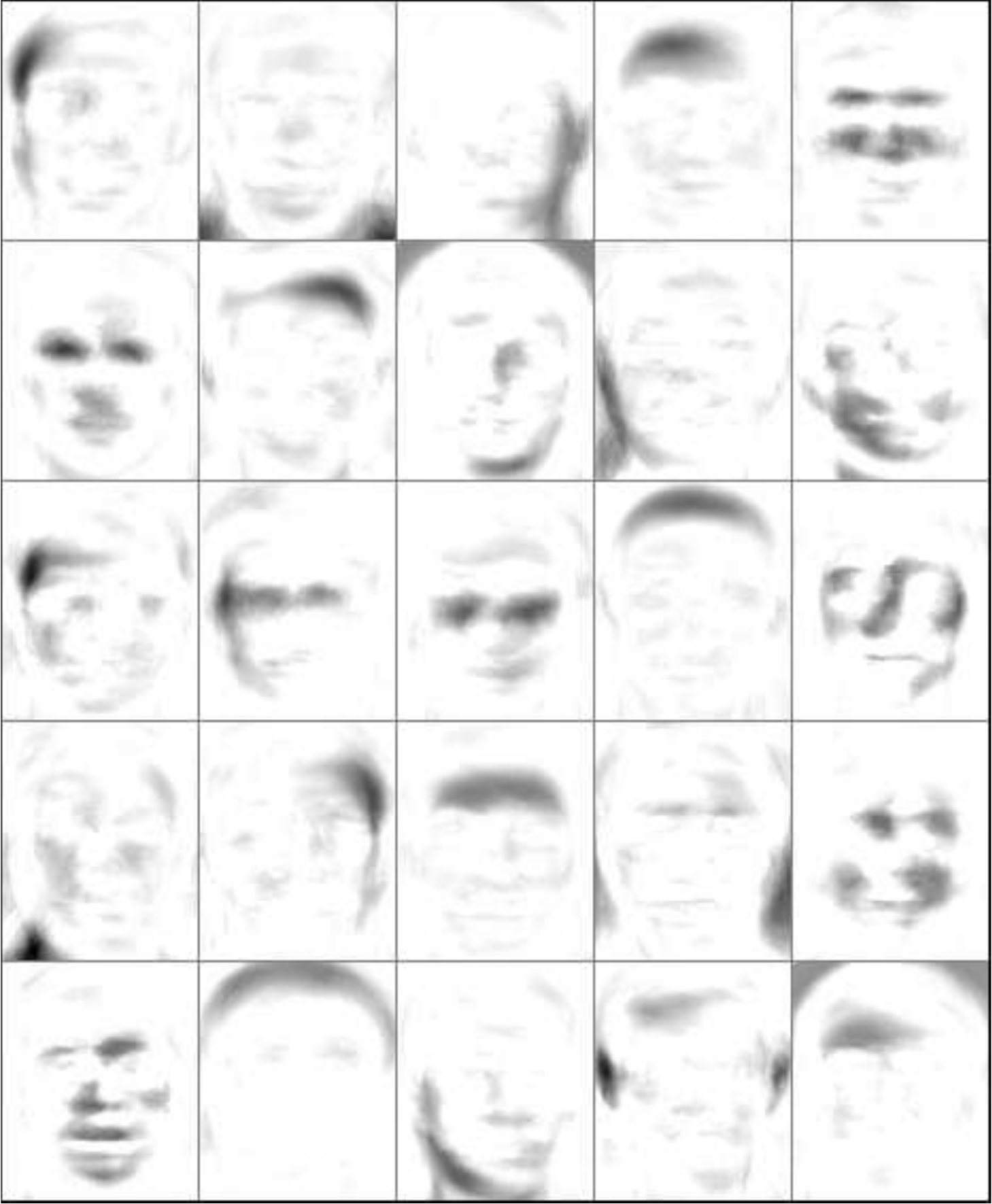}}}
   \hfill
   \subfloat[sparsity 0.75]{{\includegraphics[width=0.15\linewidth]
   {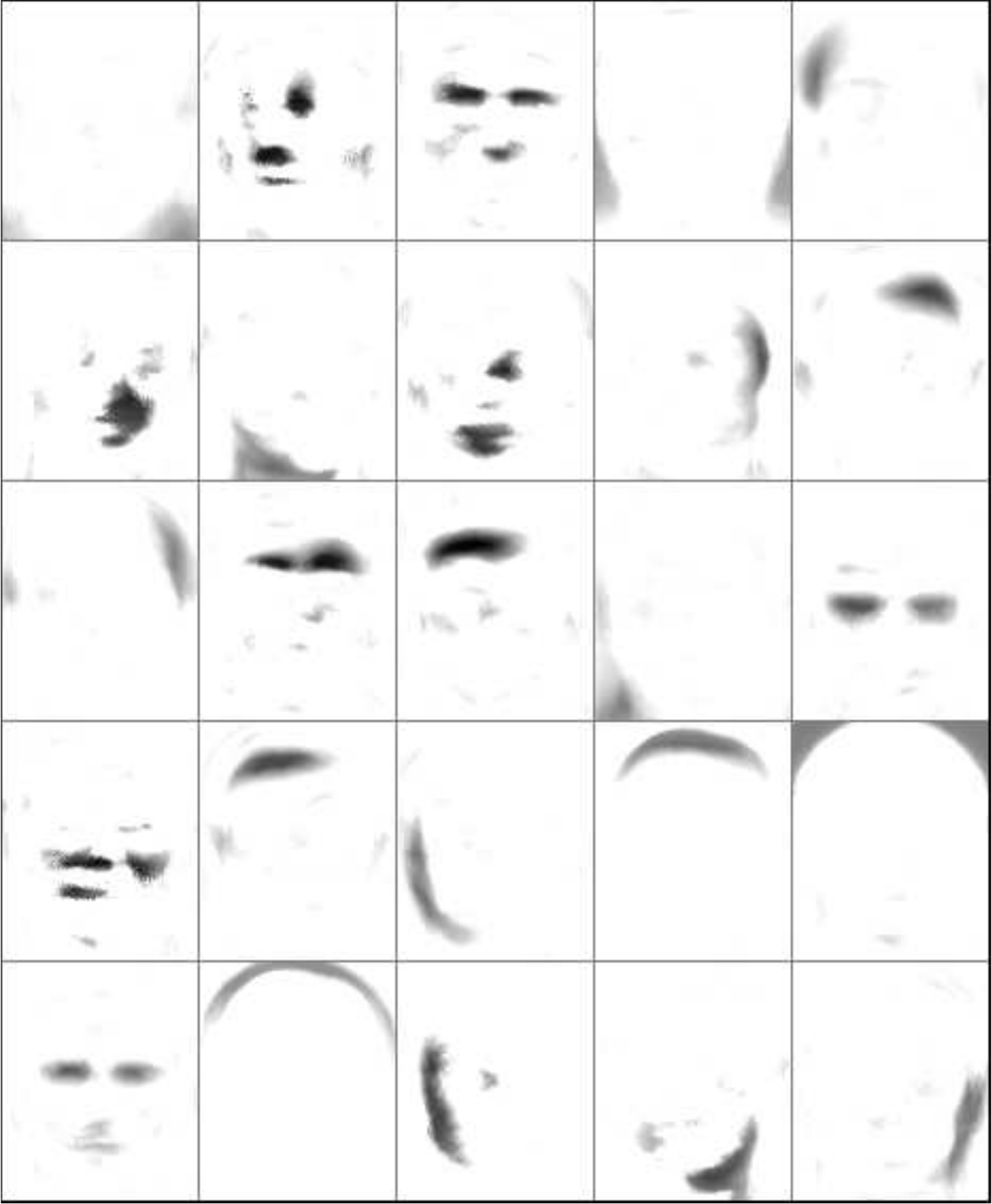}}
    {\includegraphics[width=0.15\linewidth]{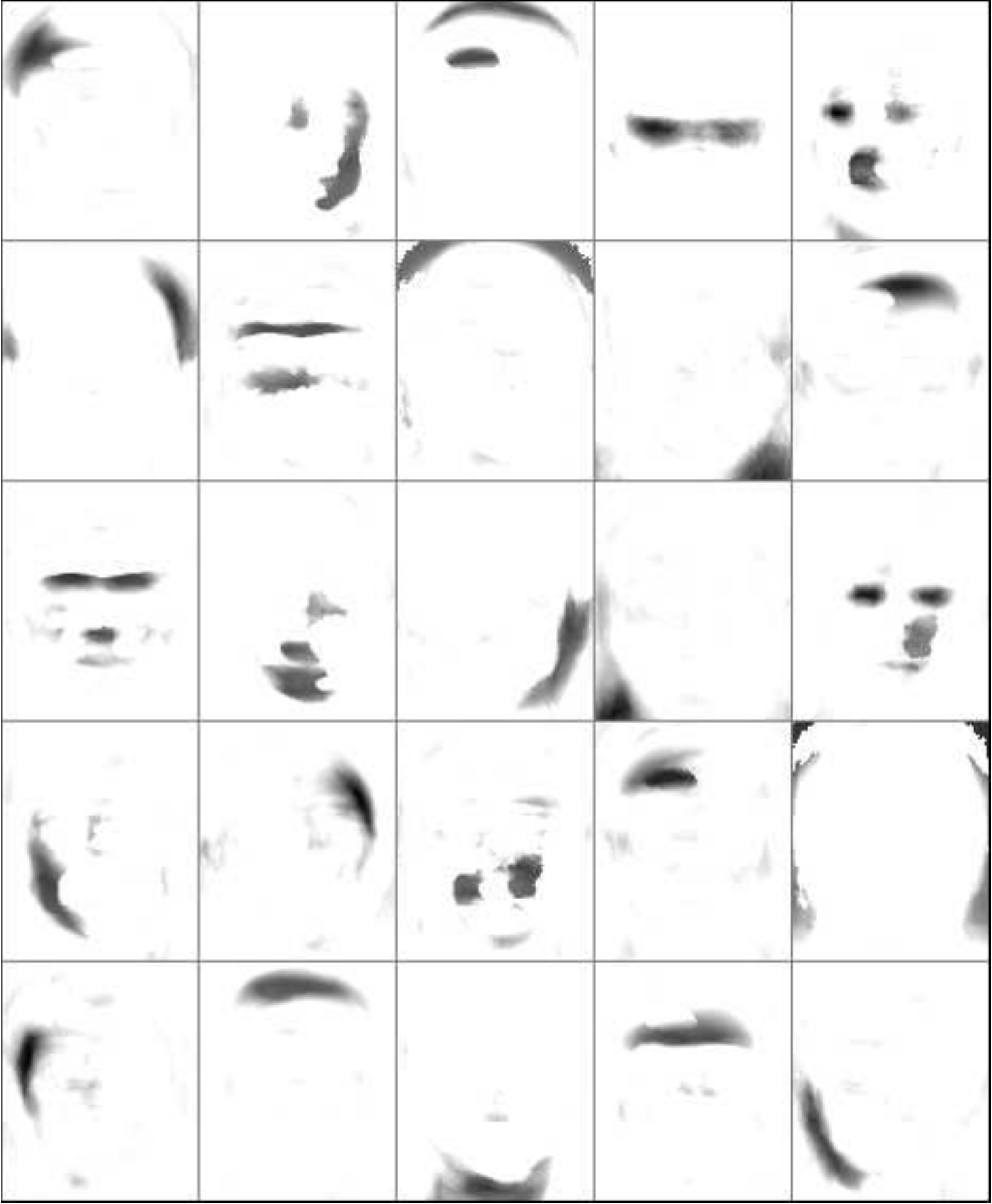}}} \\
\caption{ Feature sets from NMFSC algorithm (Left) and SSNMF algorithm
(Right) using the ORL face dataset for each sparsity value of $\alpha$ in $\{0.5,0.6,0.75\}$.
Note that SSNMF algorithm gives a parts-based representation similar
to the one recovered by NMFSC.} 
\label{fig:feat_comp}
 \end{figure*}

Finally, we note that we recover a parts-based representation as shown by
~\citeauthor{Hoyer04}~\cite{Hoyer04}. An example of the obtained features by NMFSC and ours
is shown in Figure~\ref{fig:feat_comp}.

\section{Connections to Related Work}
Other      SNMF       formulations      have      been      considered
by~\citeauthor{Hoyer2002}~\cite{Hoyer2002},~\citeauthor{Morupl0}~\cite{Morupl0}
,~\citeauthor{Kim2007}~\cite{Kim2007}, \citeauthor{Pascual06}~\cite{Pascual06}
(nsNMF) and~\citeauthor{Peharz2011}~\cite{Peharz2011}.
SNMF formulations using similar sparsity measures as used in this paper
have been considered for applications in speech and audio recordings
~\cite{weninger2012,virtanen2007}.

We note that our sparsity measure has  all  the  desirable
properties,  extensively discussed  by~\citeauthor{Hurley2009}
~\cite{Hurley2009}, except for  one (``cloning''). Cloning property
is satisfied  when two vectors of same sparsity when concatenated
maintain their sparsity value.  Dimensions in  our optimization  problem are fixed and
thus violating the cloning property is not an issue.  Compare this with the
L$_1$ norm that satisfies only one of these properties (namely ``rising
tide'').  Rising tide is the property where adding a constant to
the elements of a vector decreases the sparsity of the vector. Nevertheless, 
the  measure used  in~\citeauthor{Kim2007}  is
based  on the  L$_1$ norm.   The properties  satisfied by  the measure
in~\citeauthor{Pascual06} are  unclear because of  the implicit nature
of the sparsity formulation.

\citeauthor{Pascual06}~\cite{Pascual06}    claim     that    the    SNMF    formulation
of~\citeauthor{Hoyer04},  as   given  by  problem~\eqref{prob:sparse}
does  not capture  the  variance in  the data.  However,  some transformation of the
sparsity values is required to properly compare the two formulations~\cite{Hoyer04,
Pascual06}. Preliminary results show that the formulation given by~\citeauthor{Hoyer04}
~\cite{Hoyer04} is able to capture the variance in the data if the sparsity parameters are set 
appropriately.
~\citeauthor{Peharz2011}~\cite{Peharz2011} propose to tackle the  L$_0$ norm  
constrained NMF directly  by  projecting  from  intermediate
unconstrained solutions to the  required L$_0$ constraint.  This leads
to the  well-known problem of  getting stuck in local  minima. Indeed,
the authors re-initialize their feature  matrix with an NNLS solver to
recover from the local suboptimum.  Our formulation avoids the local
minima associated with L$_0$ norm by using a smooth surrogate.

\section{Conclusions}
\label{sec:conc}
We have proposed a new efficient algorithm to solve the 
sparse NMF problem. Experiments demonstrate the effectiveness
of our approach on real datasets of practical interest. Our algorithm
is faster over a range of sparsity values and generally performs better when
the sparsity is higher.
The speed up is mainly because of the sequential nature of the updates
in contrast to the previously employed batch updates of~\citeauthor{Hoyer04}.
Also, we presented an exact and efficient algorithm to
solve the problem of maximizing a linear objective with a sparsity constraint, which is an 
improvement over the heuristic approach in~\citeauthor{Hoyer04}.

Our approach can  be extended to other NMF  variants~\cite{Hoyer2002}.
Another  possible  application is  the  sparse  version of  nonnegative
tensor  factorization.  A  different  research direction  would be  to
scale    our     algorithm    to    handle     large    datasets   by
chunking~\cite{mairal10} and/or take advantage of distributed/parallel computational
settings~\cite{pcd11}.

\section*{Acknowledgement}

The first author
would  like to  acknowledge the  support from  NIBIB grants  1  R01 EB
000840 and 1  R01 EB 005846.  The second author  was supported by NIMH
grant 1 R01 MH076282-01. The latter  two grants were funded as part of
the  NSF/NIH  Collaborative  Research  in  Computational  Neuroscience
Program.

\bibliographystyle{plainnat}
\bibliography{papers}

\begin{thebibliography}{31}
\providecommand{\natexlab}[1]{#1}
\providecommand{\url}[1]{\texttt{#1}}
\expandafter\ifx\csname urlstyle\endcsname\relax
  \providecommand{\doi}[1]{doi: #1}\else
  \providecommand{\doi}{doi: \begingroup \urlstyle{rm}\Url}\fi

\bibitem[Arora et~al.(2012)Arora, Ge, Kannan, and Moitra]{Arora2012}
Sanjeev Arora, Rong Ge, Ravindran Kannan, and Ankur Moitra.
\newblock Computing a nonnegative matrix factorization -- provably.
\newblock In \emph{Proceedings of the 44th symposium on Theory of Computing},
  STOC '12, pages 145--162, New York, NY, USA, 2012. ACM.

\bibitem[Berry et~al.(2007)Berry, Browne, Langville, Pauca, and
  Plemmons]{berry2007}
Michael~W Berry, Murray Browne, Amy~N Langville, V~Paul Pauca, and Robert~J
  Plemmons.
\newblock Algorithms and applications for approximate nonnegative matrix
  factorization.
\newblock \emph{Computational Statistics \& Data Analysis}, 52\penalty0
  (1):\penalty0 155--173, 2007.

\bibitem[Bradley et~al.(2011)Bradley, Kyrola, Bickson, and Guestrin]{pcd11}
Joseph~K. Bradley, Aapo Kyrola, Danny Bickson, and Carlos Guestrin.
\newblock Parallel coordinate descent for {L1}-regularized loss minimization.
\newblock In \emph{ICML}, pages 321--328, 2011.

\bibitem[Buchsbaum and Bloch(2002)]{buchsbaum2002}
G.~Buchsbaum and O.~Bloch.
\newblock Color categories revealed by non-negative matrix factorization of
  munsell color spectra.
\newblock \emph{Vision research}, 42\penalty0 (5):\penalty0 559--563, 2002.

\bibitem[Chen and Ye(2011)]{chen2011}
Yunmei Chen and Xiaojing Ye.
\newblock Projection onto a simplex.
\newblock \emph{arXiv preprint arXiv:1101.6081}, 2011.

\bibitem[Cichocki and Phan(2009)]{cichockifast09}
A.~Cichocki and A.~H. Phan.
\newblock Fast local algorithms for large scale nonnegative matrix and tensor
  factorizations.
\newblock \emph{IEICE Transactions on Fundamentals of Electronics},
  92:\penalty0 708--721, 2009.

\bibitem[Cohen and Rothblum(1993)]{cohen1993}
J.~E. Cohen and U.~G. Rothblum.
\newblock Nonnegative ranks, decompositions, and factorizations of nonnegative
  matrices.
\newblock \emph{Linear Algebra and its Applications}, 190:\penalty0 149--168,
  1993.

\bibitem[Ding et~al.(2006)Ding, Li, Peng, and Park]{Ding2006}
Chris Ding, Tao Li, Wei Peng, and Haesun Park.
\newblock Orthogonal nonnegative matrix t-factorizations for clustering.
\newblock In \emph{Proceedings of the 12th ACM SIGKDD international conference
  on Knowledge discovery and data mining}, KDD '06, pages 126--135, New York,
  NY, USA, 2006. ACM.

\bibitem[Donoho and Stodden(2004)]{NIPS2003_LT10}
David Donoho and Victoria Stodden.
\newblock When does non-negative matrix factorization give a correct
  decomposition into parts?
\newblock In Sebastian Thrun, Lawrence Saul, and Bernhard Sch{\"o}lkopf,
  editors, \emph{Advances in Neural Information Processing Systems 16}. MIT
  Press, Cambridge, MA, 2004.

\bibitem[Duchi et~al.(2008)Duchi, Shalev-Shwartz, Singer, and
  Chandra]{duchi2008}
John Duchi, Shai Shalev-Shwartz, Yoram Singer, and Tushar Chandra.
\newblock Efficient projections onto the l1-ball for learning in high
  dimensions.
\newblock In \emph{Proceedings of the 25th international conference on Machine
  learning}, pages 272--279, 2008.

\bibitem[Heiler and Schn{\"o}rr(2006)]{Heiler06}
Matthias Heiler and Christoph Schn{\"o}rr.
\newblock Learning sparse representations by non-negative matrix factorization
  and sequential cone programming.
\newblock \emph{The Journal of Machine Learning Research}, 7:\penalty0 2006,
  2006.

\bibitem[Hofmann(2001)]{hofmann2001}
T.~Hofmann.
\newblock Unsupervised learning by probabilistic latent semantic analysis.
\newblock \emph{Machine Learning}, 42\penalty0 (1):\penalty0 177--196, 2001.

\bibitem[Hoyer(2002)]{Hoyer2002}
P.~O. Hoyer.
\newblock Non-negative sparse coding.
\newblock In \emph{Neural Networks for Signal Processing, 2002. Proceedings of
  the 2002 12th IEEE Workshop on}, pages 557--565, 2002.

\bibitem[Hoyer(2004)]{Hoyer04}
Patrik~O. Hoyer.
\newblock Non-negative matrix factorization with sparseness constraints.
\newblock \emph{J. Mach. Learn. Res.}, 5:\penalty0 1457--1469, December 2004.

\bibitem[Hsieh and Dhillon(2011)]{Hsieh11}
C.~J. Hsieh and I.~Dhillon.
\newblock Fast coordinate descent methods with variable selection for
  non-negative matrix factorization.
\newblock \emph{ACM SIGKDD Internation Conference on Knowledge Discovery and
  Data Mining}, page~xx, 2011.

\bibitem[Hurley and Rickard(2009)]{Hurley2009}
Niall Hurley and Scott Rickard.
\newblock Comparing measures of sparsity.
\newblock \emph{IEEE Trans. Inf. Theor.}, 55:\penalty0 4723--4741, October
  2009.

\bibitem[Kim and Park(2007)]{Kim2007}
Hyunsoo Kim and Haesun Park.
\newblock {Sparse non-negative matrix factorizations via alternating
  non-negativity-constrained least squares for microarray data analysis}.
\newblock \emph{Bioinformatics}, 23\penalty0 (12):\penalty0 1495--1502, 2007.

\bibitem[Kim and Park(2008)]{Kim2008}
Jingu Kim and Haesun Park.
\newblock Toward faster nonnegative matrix factorization: {A} new algorithm and
  comparisons.
\newblock \emph{Data Mining, IEEE International Conference on}, 0:\penalty0
  353--362, 2008.

\bibitem[Lawton and Sylvestre(1971)]{lawton1971}
W.~H. Lawton and E.~A. Sylvestre.
\newblock Self modeling curve resolution.
\newblock \emph{Technometrics}, pages 617--633, 1971.

\bibitem[Lee and Seung(1999)]{LeeSeung99}
D.~D. Lee and H.~S. Seung.
\newblock Learning the parts of objects by non-negative matrix factorization.
\newblock \emph{Nature}, 401\penalty0 (6755):\penalty0 788--791, October 1999.

\bibitem[Lee and Seung(2000)]{LeeSeung2001}
Daniel~D. Lee and Sebastian~H. Seung.
\newblock Algorithms for non-negative matrix factorization.
\newblock In \emph{NIPS}, pages 556--562, 2000.

\bibitem[Lin(2007)]{Lin2007}
Chih-Jen Lin.
\newblock Projected gradient methods for nonnegative matrix factorization.
\newblock \emph{Neural Comp.}, 19\penalty0 (10):\penalty0 2756--2779, October
  2007.

\bibitem[Mairal et~al.(2010)Mairal, Bach, Ponce, and Sapiro]{mairal10}
Julien Mairal, Francis Bach, Jean Ponce, and Guillermo Sapiro.
\newblock Online learning for matrix factorization and sparse coding.
\newblock \emph{The Journal of Machine Learning Research}, 11:\penalty0 19--60,
  2010.

\bibitem[M{\o}rup et~al.(2008)M{\o}rup, Madsen, and Hansen]{Morupl0}
Morten M{\o}rup, Kristoffer~Hougaard Madsen, and Lars~Kai Hansen.
\newblock Approximate {L0} constrained non-negative matrix and tensor
  factorization.
\newblock In \emph{ISCAS}, pages 1328--1331, 2008.

\bibitem[Pascual-Montano et~al.(2006)Pascual-Montano, Carazo, Kochi, Lehmann,
  and Pascual-Marqui]{Pascual06}
A.~Pascual-Montano, J.~M. Carazo, K.~Kochi, D.~Lehmann, and R.~D.
  Pascual-Marqui.
\newblock Nonsmooth nonnegative matrix factorization (ns{NMF}).
\newblock \emph{Pattern Analysis and Machine Intelligence, IEEE Transactions
  on}, 28\penalty0 (3):\penalty0 403--415, March 2006.

\bibitem[Peharz and Pernkopf(2011)]{Peharz2011}
R.~Peharz and F.~Pernkopf.
\newblock Sparse nonnegative matrix factorization with $l^0$-constraints.
\newblock \emph{Neurocomputing}, 2011.

\bibitem[Schmidt and Olsson(2006)]{Schmidt2006}
M.~N. Schmidt and R.~K. Olsson.
\newblock Single-channel speech separation using sparse non-negative matrix
  factorization.
\newblock In \emph{International Conference on Spoken Language Processing
  (INTERSPEECH)}, volume~2, page~1. Citeseer, 2006.

\bibitem[Theis et~al.(2005)Theis, Stadlthanner, and Tanaka]{theis2005}
Fabian~J Theis, Kurt Stadlthanner, and Toshihisa Tanaka.
\newblock First results on uniqueness of sparse non-negative matrix
  factorization.
\newblock In \emph{Proceedings of the 13th European Signal Processing
  Conference (EUSIPCO’05)}, 2005.

\bibitem[Virtanen(2007)]{virtanen2007}
Tuomas Virtanen.
\newblock Monaural sound source separation by nonnegative matrix factorization
  with temporal continuity and sparseness criteria.
\newblock \emph{Audio, Speech, and Language Processing, IEEE Transactions on},
  15\penalty0 (3):\penalty0 1066--1074, 2007.

\bibitem[Weninger et~al.(2012)Weninger, Feliu, and Schuller]{weninger2012}
Felix Weninger, Jordi Feliu, and Bjorn Schuller.
\newblock Supervised and semi-supervised suppression of background music in
  monaural speech recordings.
\newblock In \emph{Acoustics, Speech and Signal Processing (ICASSP), 2012 IEEE
  International Conference on}, pages 61--64. IEEE, 2012.

\bibitem[Xu et~al.(2003)Xu, Liu, and Gong]{Xu2003}
W.~Xu, X.~Liu, and Y.~Gong.
\newblock Document clustering based on non-negative matrix factorization.
\newblock In \emph{Proceedings of the 26th annual international ACM SIGIR
  conference on Research and development in informaion retrieval}, pages
  267--273. ACM, 2003.

\end{thebibliography}

\section*{Appendix}

\subsection*{Bi-Sparse NMF}
\label{sec:ext}
In some applications, it is desirable to set the sparsity on both matrix factors.
However, this can lead to the situation where the variance in the data is poorly
 captured~\cite{Pascual06}. To ameliorate this condition, we
formulate it as the following optimization problem and call it as
Bi-Sparse NMF:
\begin{align}
\label{prob:sparse_new}
\nonumber
  \min_{\mm{W},\mm{H},\mm{D}}\frac{1}{2}\|\mm{X}&-\mm{W}\mm{D}\mm{H}\|_F^2  \\ 
\nonumber
                 & \textrm{s.t.} \mm{W}\ge\mm{0},\mm{H}\ge\mm{0},\mm{D}\ge\mm{0}\\
\nonumber
  &  \|\mm{W}_j\|_2 =1,  \textrm{sp}(\mm{W}_j) =\alpha, \forall j\in\{1,\cdots,r\} \\
  &  \|\mm{H}^i\|_2 =1, \textrm{sp}(\mm{H}^i) =\beta, \forall i\in\{1,\cdots,r\} 
\end{align}
where $\mm{D}$ is a $r \times r$ matrix. In the above formulation, we 
constrain the L$_2$ norms of the columns of matrix $\mm{W}$ to unity. 
Similarly, we  constrain the L$_2$ norms of rows of matrix $\mm{H}$ to be unity.
This scaling is absorbed by the matrix $\mm{D}$. Note that this formulation
with the matrix $\mm{D}$ constrained to be diagonal 
is equivalent to the one proposed in~\citeauthor{Hoyer04} when both the matrix factors
have their sparsity specified. 

We can solve for the matrix $\mm{D}$ with any NNLS solver. 
A concrete algorithm is the one presented in~\citeauthor{Ding2006} and is
reproduced here for convenience (Algorithm~\ref{alg:Ding}).
If $\mm{D}$ is a diagonal matrix, we only update 
the diagonal terms and maintain the rest at zero.
Algorithms~\ref{alg:leeseung} and~\ref{alg:Ding}  can be sped up
by  pre-computing the matrix products which are unchanged during the iterations.
\begin{algorithm}
\caption{ $\textrm{Diag-mult}(\mm{X},\mm{W},\mm{H},\mm{D})$}
\label{alg:Ding}
\begin{algorithmic}
\REPEAT
\STATE 
$\mm{D} = \mm{D} \odot \frac{\trans{\mm{W}}\mm{X}\mm{H}}{\trans{\mm{W}}\mm{W}\mm{D}\mm{H}
\trans{\mm{H}}}$
\UNTIL convergence
\STATE Output: Matrix $\mm{D}$.
\end{algorithmic}
\end{algorithm}

Also, the matrix $\mm{D}$ captures the variance
of the dataset when we have sparsity set on both the matrices $\mm{W},\mm{H}$. 

\end{document}